\documentclass[10pt]{amsart}

\usepackage{float}
\usepackage[utf8]{inputenc}
\usepackage{kantlipsum}
\usepackage[ colorlinks=true,linkcolor=blue, citecolor={green}]{hyperref}
\usepackage{lmodern}\usepackage[T1]{fontenc}
\usepackage{amsmath,amssymb,amsfonts,euscript,graphicx,hyperref,indentfirst}
\usepackage{enumerate,linegoal}
\usepackage[all]{xy}
\usepackage{tikz-cd}
\usepackage{stmaryrd}
\usepackage{relsize,exscale}
\usepackage{bigints}
\usepackage{cancel}
\usepackage{accents}
\usepackage{bbm}
\usepackage[font=footnotesize]{caption} 
\usepackage{diagbox}
\usepackage{soul}
\usepackage[extra]{tipa}
\usepackage{array, boldline, makecell, booktabs}
\usepackage{multirow}
\usepackage{algpseudocode}
\usepackage{mathrsfs}
\usepackage{subcaption}
\usepackage{colortbl}

\usepackage{xcolor}

\newcolumntype{?}{!{\vrule width 2pt}}
\newcolumntype{a}{!{\vrule width 1pt}}

\calclayout

\newtheorem*{corollary*}{Corollary}

\newtheorem{definition-theorem}[theorem]{Definition-Theorem}

\theoremstyle{definition}

\newtheorem{definition-notation}[theorem]{Definition-Notation}

\theoremstyle{remark}

\makeatletter
\def\l@subsection{\@tocline{2}{0pt}{2.5pc}{5pc}{}} 
\makeatother

\numberwithin{equation}{section}

\title[Mech. Interp. for LLMs with applications to Financial Services]{Mechanistic interpretability of large language models with applications to the financial services industry}
\author{Ashkan Golgoon$^{*,\ddagger}\quad$, Khashayar Filom$^{*,\dagger}\quad$, Arjun Ravi Kannan$^{*,\S}\quad$}
\thanks{$^*$\emph{Emerging Capabilities Research Group, Discover Financial Services Inc., Riverwoods, IL}}
\thanks{$^\ddagger$ Co-first author, \texttt{ashkangolgoon@gmail.com}}
\thanks{$^\dagger$ Co-first author, \texttt{khashayar.1367@gmail.com}}
\thanks{$^\S$ \texttt{arjun.kannan@gmail.com}}
\date{October 2024}

\begin{document}

\begin{abstract}
Large Language Models such as GPTs (Generative Pre-trained Transformers) exhibit remarkable capabilities across a broad spectrum of applications. Nevertheless, due to their intrinsic complexity, these models present substantial challenges in interpreting their internal decision-making processes. This lack of transparency poses critical challenges when it comes to their adaptation by financial institutions, where concerns and accountability regarding bias, fairness, and reliability are of paramount importance. Mechanistic interpretability aims at reverse engineering complex AI models such as transformers. In this paper, we are pioneering the use of mechanistic interpretability to shed some light on the inner workings of large language models for use in financial services applications. We offer several examples of how algorithmic tasks can be designed for compliance monitoring purposes. In particular, we investigate GPT-2 Small's attention pattern when prompted to identify potential violation of Fair Lending laws. Using direct logit attribution, we study the contributions of each layer and its corresponding attention heads to the logit difference in the residual stream. Finally, we design clean and corrupted prompts and use activation patching as a causal intervention method to localize our task completion components further. We observe that the (positive) heads $10.2$ (head $2$, layer $10$), $10.7$, and $11.3$, as well as the (negative) heads $9.6$ and $10.6$ play a significant role in the task completion.
\end{abstract}

\maketitle

\begin{description}
\item[Keywords:] Mechanistic Interpretability, Large Language Models (LLMs), Transformer Circuits, \newline FinTech, Natural Language Processing.
\end{description}

The goal of this paper is to introduce the reader to the emerging field of \textit{mechanistic interpretability} and its potential applications to financial services, especially when it comes to understanding the inner workings of 
\textit{Large Language Models} (LLMs) in financial services. 

\section{Background on LLMs}
The release of ChatGPT by OpenAI in late 2022 stunned the world as the chatbot set new milestones surpassing all previous publicly available systems and triggered discussions about potential risks \cite{Reuters,NYT2,NYT1}. At the heart of the current large language model (LLM) revolution lies the \textit{transformer architecture}. Below, we provide a short introduction on transformers, followed by LLMs.  

\begin{figure}
\includegraphics[width=8cm]{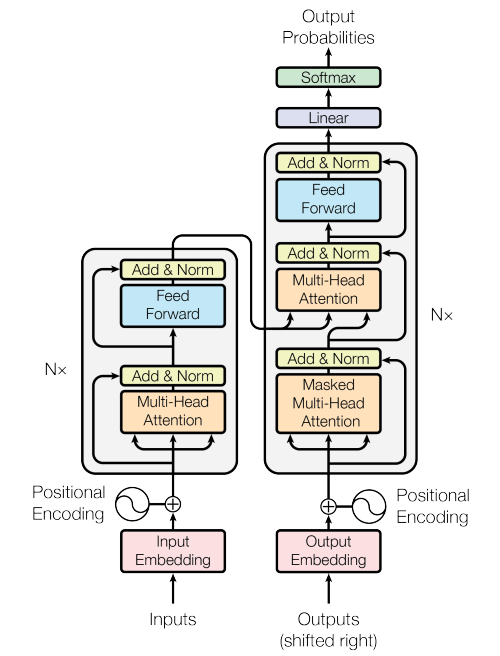}
\caption{The transformer architecture. Picture adapted from \cite{vaswani2017attention}.}
\label{fig:transformer}
\end{figure}

\subsection{The transformer architecture}
Transformers are deep feed-forward neural networks that leverage the \textit{attention mechanism}. 
They excel in sequence modeling tasks, especially in natural language processing (NLP) \cite{2022arXiv220709238P}.
Before the advent of the transformer architecture in the landmark paper \cite{vaswani2017attention}, recurrent neural network (RNN) architectures such as \textit{Long-Short Term Memory} (LSTM) were common for NLP and sequence modeling tasks. Such models rely on an internal hidden state and must process the data sequentially. 
On the other hand, transformers, which are based on the attention mechanism, are superior in capturing long-term dependencies. This is due to the non-sequential and parallel manner by which they process the data. Transformers, moreover, allow \textit{transfer learning}---they can be pre-trained on large corpora and then \textit{fine-tuned} for downstream tasks---a fact that  facilitates developing new AI applications tailored for in-house datasets based on the existing \textit{foundation models} such as BERT or GPT-4.

In the context of sequential data, the goal can be to learn the probability distribution of the next token in \textit{sequence modeling tasks} (e.g., language modeling), the probability distribution of a sequence conditioned on another sequence in \textit{sequence-to-sequence tasks} (e.g., machine translation), or the probability score in a \textit{text classification task} (e.g., sentiment analysis). As alluded to earlier, the transformer architecture shines in such tasks via utilizing the attention mechanism. The distinctions made above between different tasks are reflected in the attention type---\textit{bidirectional}/\textit{unmasked} or \textit{unidirectional}/\textit{masked} attention---and the presence or absence of \textit{encoder} and \textit{decoder} stacks  in the network \cite{2022arXiv220709238P}. The difference between these two stacks is that the former maps into a latent space while the latter takes its inputs from a latent space. Figure \ref{fig:transformer}, from the original paper on transformers \cite{vaswani2017attention}, illustrates an encoder-decoder transformer architecture. Below, we briefly discuss the attention mechanism, followed by some other components appearing in that illustration, e.g., a \textit{tokenization} step in the context of language tasks. 

The attention mechanism is based on \textit{key}, \textit{query} and \textit{value} vectors---which are all learnable. In its simplest form, the current token, the one to be predicted, is mapped to a query vector $\mathbf{q}$; and the tokens in the context are mapped to key vectors $\mathbf{k}_t$ and value vectors $\mathbf{v}_t$ (as $t$ varies, different tokens in the context are captured). Key and query vectors are of the same dimension, say $d_{\rm{attn}}$, while the dimension of the value vectors may be different. Indeed, that dimension coincides with the dimension of the output vector 
(a representation of token and context combined) because the output is a linear combination of value vectors $\mathbf{v}_t$. The coefficients of this  combination are the entries of 
${\rm{softmax}}\Big(\frac{\mathbf{q}^{\rm{T}}K}{\sqrt{d_{\rm{attn}}}}\Big)$
where the softmax function is applied to a normalization of a matrix product involving the query vector and the matrix $K$ formed by the key vectors $\mathbf{k}_t$. The attention mechanism just described was based on a single query; that is, we outlined a single \textit{attention head}. In practice, transformers use a \textit{multi-head attention} mechanism where multiple attention heads are run in parallel and their outputs are combined by concatenation and then projection. We refer the reader to \cite{vaswani2017attention} for more details, or to \cite{penke2022mathematician,2022arXiv220709238P} for mathematically rigorous treatments of the attention mechanism. 

We end this subsection by briefly mentioning some of the other components of a transformer model (cf. Figure \ref{fig:transformer}). We follow \cite{2022arXiv220709238P} where precise pseudocodes for these components are presented: 
\begin{itemize}
\item[$\blacktriangleright$] \textit{Token Embedding}) A vector representation of each vocabulary element (token) is learned. 
\item[$\blacktriangleright$] \textit{Positional Embedding}) As a remedy to the lack of recurrence or convolution in transformers, it is suggested to inject information about the position of tokens via adding certain sinusoidal terms to input embeddings \cite{vaswani2017attention}.
\item[$\blacktriangleright$] \textit{MLP}) Blocks of fully-connected feed-forward neural networks (multi-layer perceptrons) are occasionally used in a transformer. 
\item[$\blacktriangleright$] \textit{Add \& Norm Layers}) \textit{Residual connections} and \textit{layer normalization} are used to help with the vanishing gradient problem during training, and to make the training faster and more stable.
\item[$\blacktriangleright$] \textit{Unembedding}) The model learns to convert vector representations of tokens and their contexts to a distribution over vocabulary elements. 
\end{itemize}

\subsection{Large Language Models (LLMs)}
Recent transformer-based LLMs are systems pre-trained on an enormous amount of data with an astronomical numbers of parameters. The table below, adapted from survey \cite{2024arXiv240206196M}, summarizes certain aspects of famous LLMs as of 2024, e.g., their type, number of parameters, number of tokens etc. 
We refer the interested reader to lecture notes \cite{2023arXiv230705782D} for an introduction to LLMs. The exposition is aimed at mathematicians and physicists, and discusses the historical pretext and the phenomenology of language models among other topics. 

The advent of LLMs has stimulated conversations and posed urgent questions about this disruptive technology, including on the \textit{emergent} properties of LLMs \cite{wei2022emergent}, on their \textit{alignment} with human values \cite{shen2023large}, on the \textit{bias} present in their training data \cite{li2023survey}, on the \textit{hallucination} problem \cite{ye2023cognitive}, and finally, on the challenge of interpreting LLMs \cite{singh2024rethinking}. As a matter of fact, LLMs are expected to produce a rapidly growing array of risks which makes research on safety and governance mechanisms for them even more crucial \cite{2023arXiv230400612B}. In this paper, we focus on the interpretability question, beginning with a short introduction to the field of mechanistic interpretability in the next section.

\begin{figure*}
\includegraphics[width=16cm]{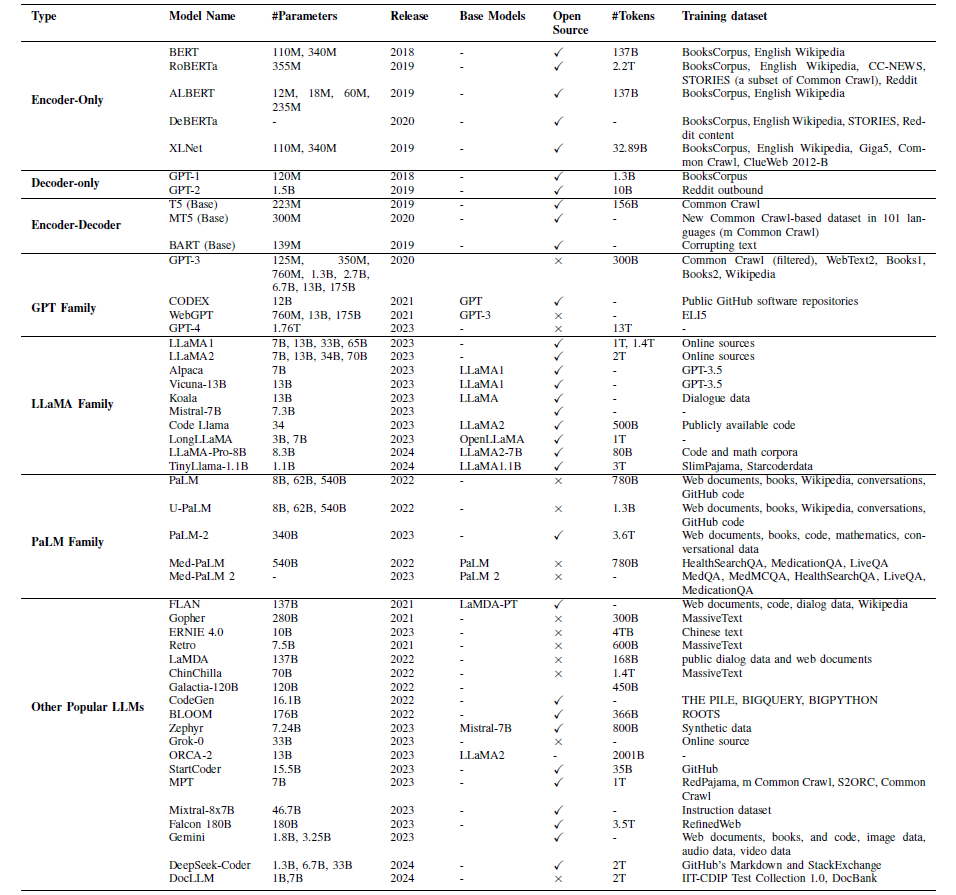}
\caption*{Table 1: An overview of popular large language models as of February 2024 (courtesy of \cite{2024arXiv240206196M}).}
\label{fig:table}
\end{figure*}

\section{Mechanistic interpretability}
\label{section2}
The nascent field of mechanistic interpretability seeks to ``reverse engineer'' neural networks. This endeavor can be construed in parallel with understanding the compiled binary program run on a virtual machine where the binary code and virtual machine/interpreter correspond to the parameters and the architecture of a neural network. In this setting, variables/memory locations roughly correspond to neurons or  other ``independent units'' a neural network representation can be decomposed into \cite{OlahIntuition}. 
For basic resources on the topic, we refer the reader to the guide \cite{NandaGuide}, the glossary \cite{NandaGloss}, the TransformerLens library \cite{nanda2022transformerlens}, or the course \cite{ARENA}.

\subsection{Motivation}
Despite their immense success in various tasks, the lack of transparency of LLMs presents challenges such as hallucination, toxicity, unfairness, and misalignment with human values which can hinder safe deployment of these models. Thus, there is an urgent need for a deeper understanding of the inner functioning of LLMs. Mechanistic interpretability is an important explanation technique used to this end. In this paradigm, one strives to understand an LLM at the level of neurons, \textit{circuits}, and attention heads, i.e., at a micro scale (as opposed to \textit{representation engineering} \cite{2024arXiv240210688Z,zou2023representation} where the explanation occurs at a macro scale). In XAI (eXplainable AI) terminology, mechanistic interpretability  can be described as a global, post-hoc, model-specific and white-box (i.e., needs access to model's internals)  approach \cite{2024arXiv240210688Z}.
Apart from helping to address societal risks, insights from mechanistic interpretability can help with ``editing'' an LLM for better performance. It can furthermore be utilized to explain training phenomena such as \textit{grokking} and \textit{memorization}; cf. \cite{2023arXiv230105217N}.

\subsection{Circuits}
\begin{figure}
\includegraphics[width=15cm]{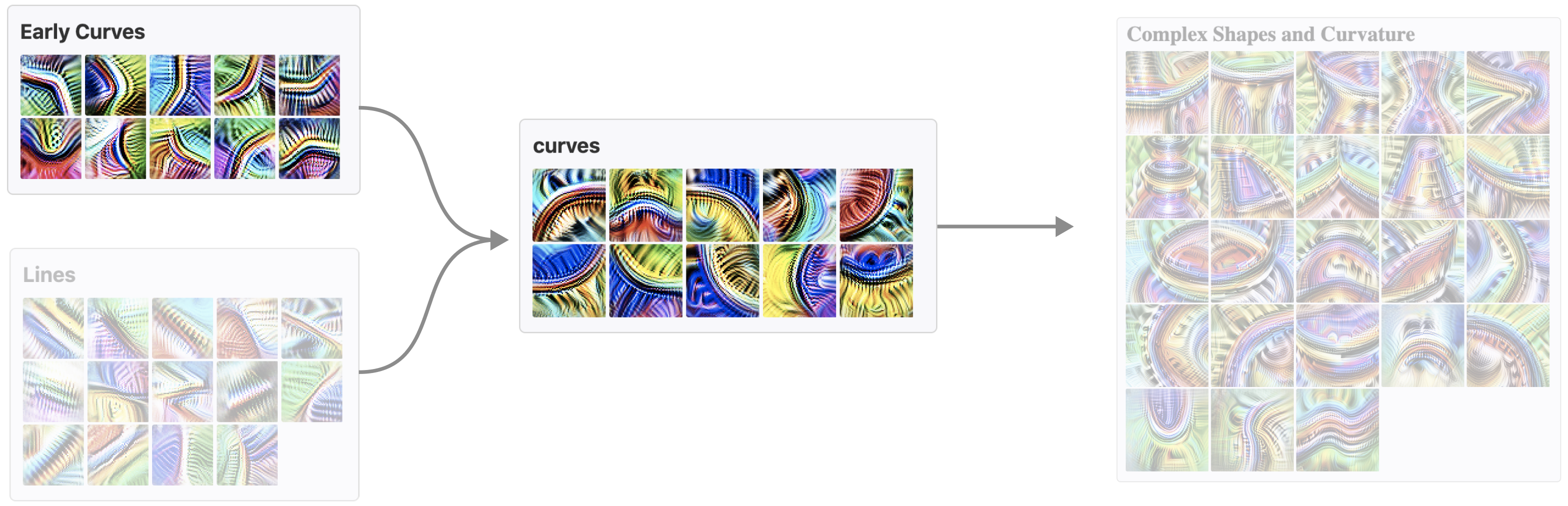}
\caption{Curve detector and line detector neurons in early layers form ``full curve detector circuits'' that can be used to create complex shapes and geometry. The picture is based on experiments with the \href{https://ieeexplore.ieee.org/document/7298594}{Inception} convolutional architecture.
It is from \cite{olah2020zoom} (taken from \url{https://github.com/distillpub/post--circuits-zoom-in} under the CC-BY 4.0 license).}
\label{fig:circuit}
\end{figure}

By analogy with cellular biology, researchers have tried to understand complicated neural networks by ``zooming in'' and investigating ``the fundamental units'' building a network, and the connections between them \cite{olah2020zoom}. 
These building blocks are called ``features''. Each feature corresponds to a ``direction'', meaning a vector in the representation of a layer of the neural net. This can be an individual neuron or a linear combination of neurons in a layer. Features are connected to each other by weights to form ``circuits''.  More precisely, a circuit is a computational subgraph of the neural network where each edge connects two neurons/directions in adjacent layers, and comes with weights which are the weights shared between them in the network. It is claimed (admittedly speculatively) in \cite{olah2020zoom} that features are typically meaningful, i.e., they correspond to articulable properties of the input; and the circuits correspond to meaningful algorithms encoded in neural networks' weights. These ideas are perhaps better understood in the context of vision models---neurons may pick certain aspects of the input image and then group together to form more sophisticated detectors; see Figure \ref{fig:circuit}. 

\begin{figure}
\includegraphics[width=15cm,height=6cm]{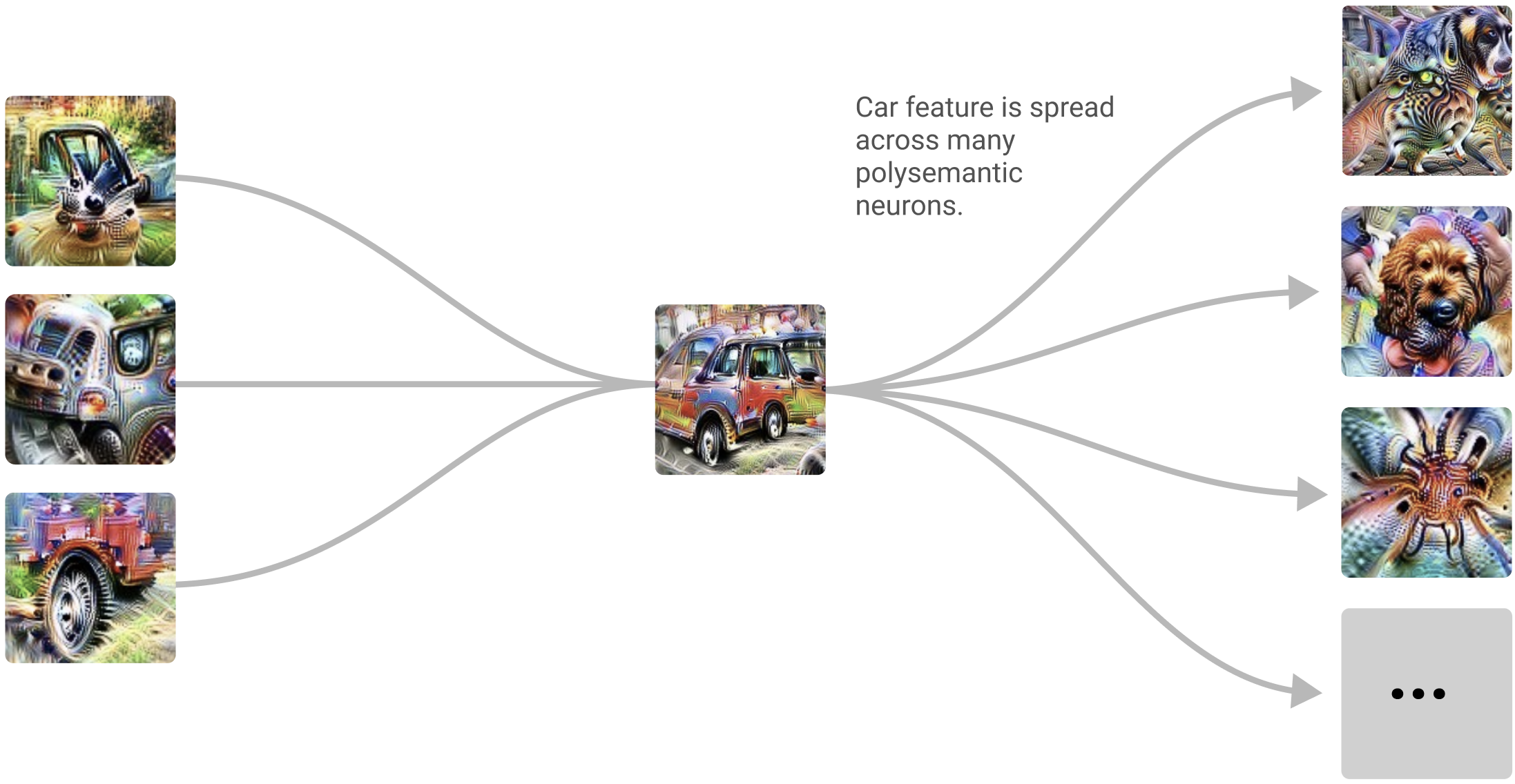}
\caption{An illustration of polysemantic neurons in the context of vision models  adapted from \cite{olah2020zoom} (taken from \url{https://github.com/distillpub/post--circuits-zoom-in} under the CC-BY 4.0 license).}
\label{fig:poly}
\end{figure}

Above, we presented a very brief account of features and circuits. Mechanistic interpretability is a rapidly growing field and many other aspects of these concepts have been studied. 
\begin{itemize}
\item An important hurdle to explaining neural networks through circuits is the existence of \textit{polysemantic} neurons; that is, neurons that respond to/get activated by multiple unrelated inputs; cf. Figure \ref{fig:poly}. Polysemanticity can be due to \textit{superposition}, meaning when a circuit spreads a feature across many neurons (which can be inevitable since there are more features than neurons); see \cite{NandaGloss,NandaGuide} for more details, and the work \cite{templeton2024scaling} on extracting \textit{monosemantic} features. 
\item \textit{Universality} (or \textit{convergent learning}) is the claim that  analogous features and circuits form across models and tasks. This hypothesis, inspired by cell theory, clearly shapes the mechanistic interpretability research by suggesting focusing on ``universal'' neurons, features or circuits. This line of research has been investigated in \cite{2023arXiv230203025C,2024arXiv240112181G}.  
\end{itemize}

\begin{figure}
\includegraphics[width=18cm]{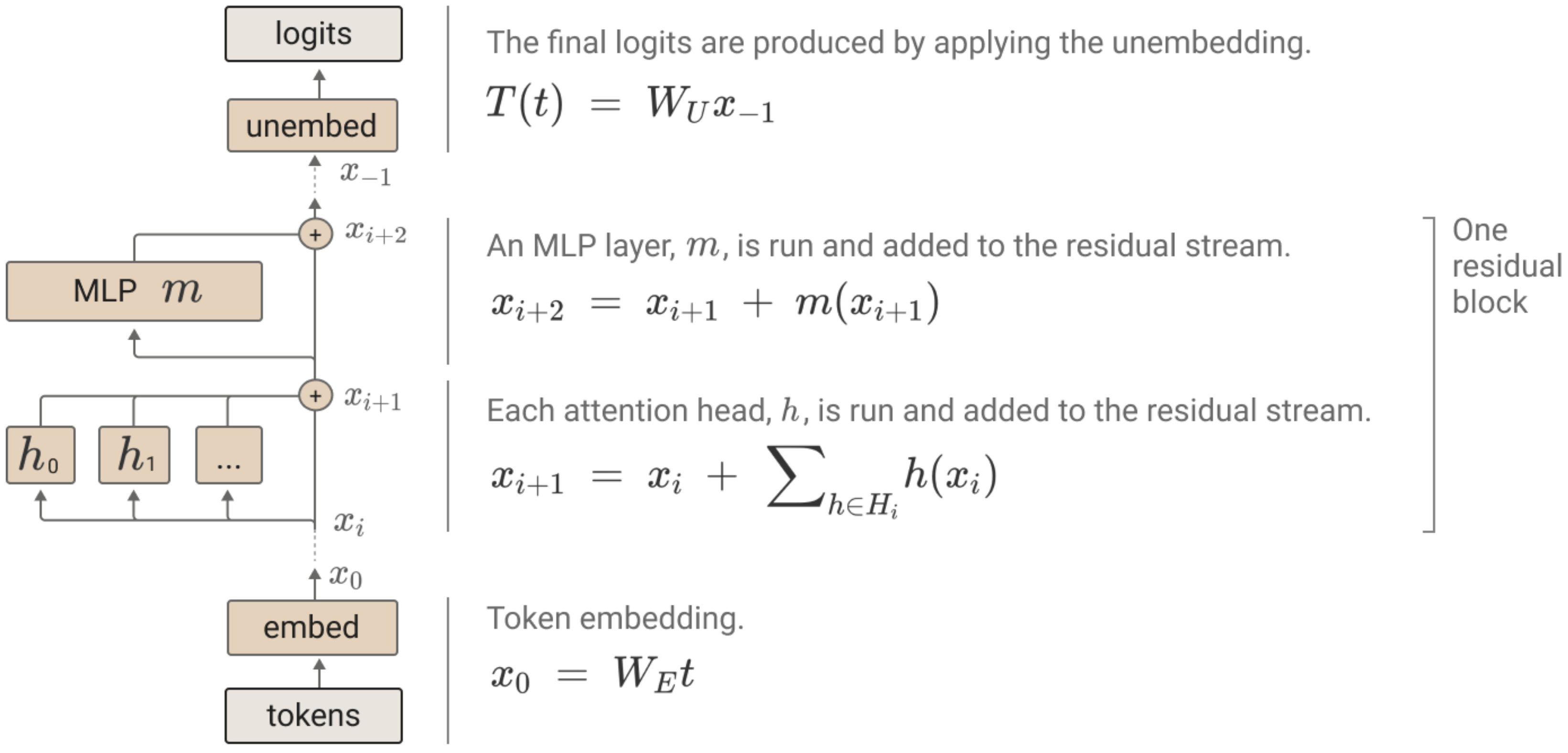}
\caption{A depiction of the residual stream in a transformer (courtesy of \cite{elhage2021mathematical}).}
\label{fig:residual}
\end{figure}

\begin{figure}
\includegraphics[width=15cm,height=7cm]{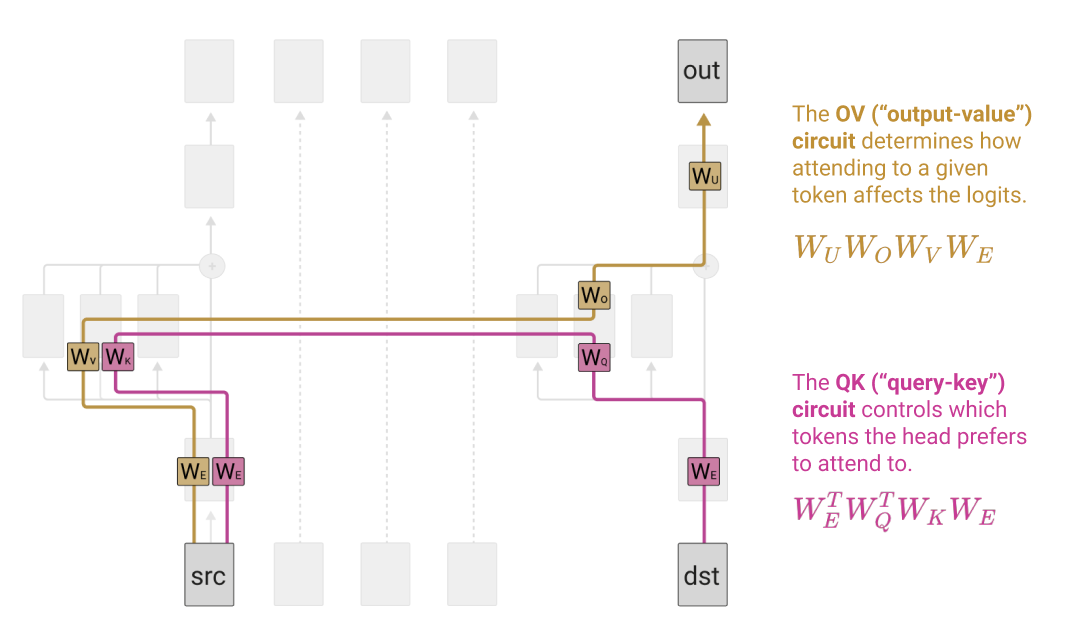}
\includegraphics[width=15cm]{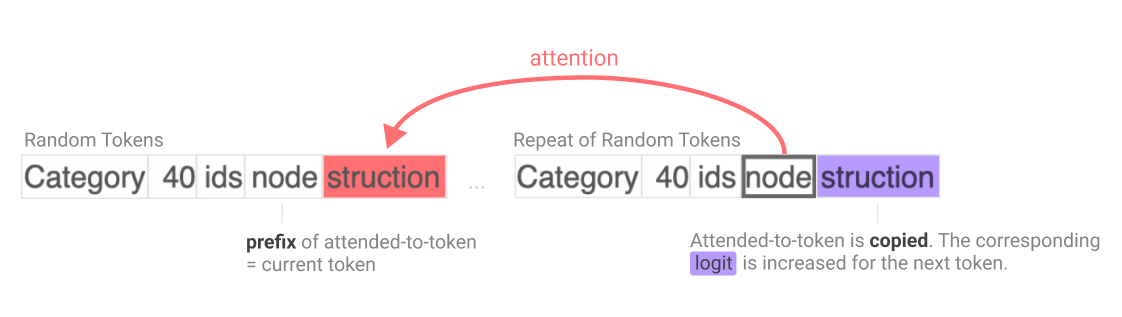}
\caption{Top, an illustration of QK (query-key) and OV (output-value) circuits (courtesy of \cite{elhage2021mathematical}). Bottom, an illustration of induction heads' behavior (courtesy of \cite{olsson2022context}).}
\label{fig:circuits}
\end{figure}

We now delve into a more precise treatment of circuits following \cite{elhage2021mathematical,ARENA}. In transformers, a layer's output is added to the \textit{residual stream}, i.e., the sum of outputs of the previous layers and the original embedding. The residual stream can be considered as a ``communication channel'' between different transformer components such as MLP layers and attention heads; see Figure \ref{fig:residual}. The function of attention heads can then be understood as moving the information: reading from the residual stream of one token, and writing to the residual stream of another token. Following \cite[chap. 1.1 \S2, chap. 1.2 \S4]{ARENA}, and \cite{elhage2021mathematical}, we argue that each attention head consists of two circuits (Figure \ref{fig:circuits} (top)):
\begin{itemize}
\item A QK (query-key) circuit ``determines where to move information to and from'' \cite[chap. 1.1 \S2]{ARENA}. For attention head $h$, this can be captured by the matrix product $W_E^{\text{\tiny T}}(W^h_Q)^{\text{\tiny T}}W^h_KW_E$.
\item An OV (output-value) circuit  ``determines what information to move'' \cite[chap. 1.1 \S2]{ARENA}. For attention head $h$, this can be captured by the matrix product $W_UW^h_OW^h_VW_E$.
\end{itemize}
In what appeared above, matrices $W_E$ and $W_U$ capture embedding and unembedding of tokens; $W^h_K$, $W^h_Q$ and $W^h_V$ denote the key, query and value matrices of the attention head $h$; and $W^{h}_O$ captures the attention head's output weights.  A thorough study of these circuits for transformers with no more than two layers and no MLP layer can be found in \cite{elhage2021mathematical}. Great insights can be gained from investigating these toy models: when there is one transformer layer, or none, 
there are concrete descriptions in terms of n-gram models, whereas two-layer transformers are much more capable since now attention heads can compose (in three different ways corresponding to keys, queries, and values), and this results in creation of \textit{induction heads} (Figure \ref{fig:circuits} (bottom)). Going beyond toy examples, paper  \cite{olsson2022context} studies induction heads for larger models: they are implemented by circuits consisting of  attention heads in different layers that work together to copy or complete patterns, and thus, contributing to \textit{in-context learning}. 

We focus on \textit{circuit discovery} in the next subsection, especially the idea of \textit{algorithmic tasks} which is essential to the financial services applications proposed in the last section.

\begin{figure}
\includegraphics[width=5cm]{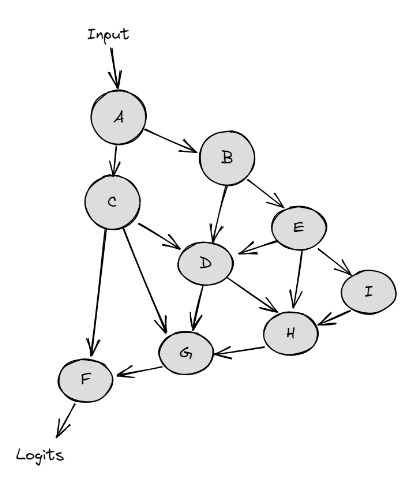}
\includegraphics[width=8cm]{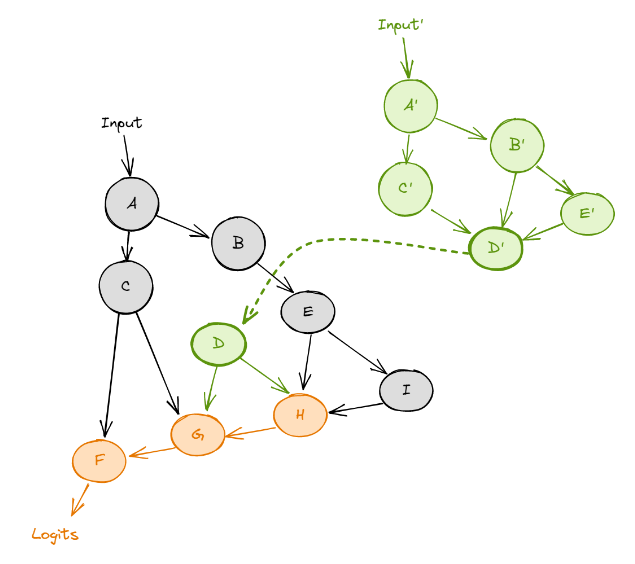}
\caption{In activation patching, a clean run (e.g., \textsf{''When Mary and John went to the store, John gave a drink to ...''} in the IOI task) is considered along with a corrupted run for which the answer is flipped (e.g., \textsf{''When Mary and John went to the store, Mary gave a drink to ...''}). The clean run is captured on the left. On the right, an activation from the corrupted run is patched in the corresponding one from the clean run. (Pictures courtesy of \cite{ARENA}.)
}
\label{fig:activation}
\end{figure}

\subsection{Circuit discovery}
\label{section23}
There is a growing body of literature on circuit discovery for transformers; cf. \cite{conmy2023towards}. As a concrete example, the phenomenon of grokking, i.e., sudden transition from overfitting to generalization after numerous training steps, can be explained through circuit formation. This is the content of paper \cite{2023arXiv230105217N} where grokking is reverse engineered for transformer models trained for simple arithmetic tasks. Also, see \cite{2024arXiv240215175H} where grokking and ``emergent'' abilities in LLMs are studied through the lens of circuits. 

For our purposes, we shall follow the approach of \cite{wang2022interpretability} which is based on algorithmic tasks, an idea that we believe carries over easily to financial services applications.
The paper focuses on the \textit{Indirect Object Identification} (IOI) language task with the GPT-2 Small model (\cite{radford2019language}), a language model comprised of 12 layers and about 80 million parameters. To give an example of this task, a sentence such as \textsf{''When Mary and John went to the store, John gave a drink to ...''} should be completed with the word \textsf{''Mary''} instead of \textsf{''John''}. In that paper, a circuit (here, meaning a computational subgraph) of GPT-2 Small, consisting of 26 attention heads, is identified as being responsible for the IOI task. These are further broken down into 7 categories each responsible for a human-understandable purpose, e.g., ``duplicate token heads'' that identify tokens already appeared in the sentence.

Algorithmic examples of the IOI instances are constructed as follows (see \cite[Appendix E]{wang2022interpretability} and \cite[IOI Dataset]{easytransformer}):
\begin{itemize}
    \item ABC-TEMPLATES: \textsf{''Afterwards [A], [B] and [C] went to the [PLACE]. [B] and [C] gave a [OBJECT] to [A]''},
    \item BABA-TEMPLATES: \textsf{''When [B] and [A] got a [OBJECT] at the [PLACE], [B] decided to give the [OBJECT] to [A]''},
    \item BABA-LONG-TEMPLATES: \textsf{''Then in the morning, [B] and [A] went to the [PLACE]. [B] gave a [OBJECT] to [A]''},
    \item  BABA-LATE-IOS: \textsf{''Afterwards, [B] and [A] went to the [PLACE]. [B] gave a [OBJECT] to [A]'',}
    \item BABA-EARLY-IOS: \textsf{''After the lunch [B] and [A] went to the [PLACE], and [B] gave a [OBJECT] to [A]''},
\end{itemize}
such that each name was drawn from a list of 100 English first names. The place and the object were chosen from a hand-made list of 20 common words. Using these templates, a dataset of samples for IOI is created. Next, GPT-2 Small's performance is evaluated on the prompted dataset and mechanistic interpretability of the model pertaining to the IOI task is investigated.

\begin{figure}
    \centering
    \includegraphics[width=10cm]{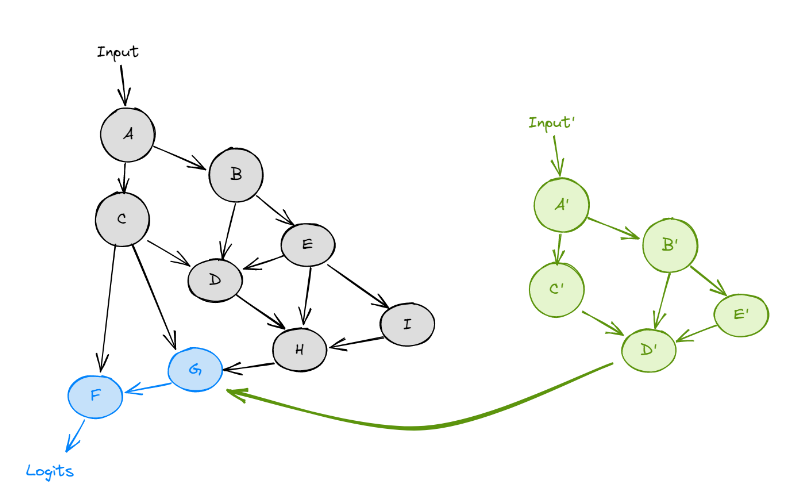}
    \caption{Thinking of nodes as attention heads and edges as representations of how two attention heads are tangled in the residual stream, the picture illustrates path patching where edges are replaced (unlike the activation patching where nodes are replaced; cf. Figure \ref{fig:activation}). (Picture courtesy of \cite{ARENA}.)}
    \label{fig:path}
\end{figure}

After this high-level overview of \cite{wang2022interpretability}, we delve into the steps and tools involved in the circuit discovery carried out therein following the exposition in \cite[chap. 1.3]{ARENA}.
\begin{itemize}
\item Recall that the probabilities that a language model outputs for predicting the next token 
are obtained from applying a softmax function to tokens' \textit{logit values}. The difference of logit values, e.g., the expression $logit$(\textsf{"Mary"}) $-$ $logit$(\textsf{"John"}) in the case of the IOI example above, can be utilized as a performance metric. 
\item The contribution of a layer (a transformer block), and the attention heads inside it, to the residual stream can be quantified by considering the logit difference after that layer as if the subsequent layers are ignored. This \textit{direct logit attribution} technique identifies heads that contribute the most to the residual stream, and can be conducted with the TransformerLens library \cite{nanda2022transformerlens}. See \cite[chap. 1.3 \S2]{ARENA} for more details.
\item A more sophisticated approach to end-to-end circuit discovery, which goes beyond just looking at what happens at the very end of a circuit, is \textit{activation patching}, as well as its more refined version \textit{path patching}. 
\begin{itemize}
    \item[$\blacktriangleright$]   Let us begin with the activation patching tool, the idea that was first introduced in \cite{meng2022locating}. Two different runs of the model are considered, a ``clean'' run and a ``corrupted'' one. The output answers for them are correct and incorrect, respectively. We intervene on a specific activation from the corrupted run by replacing it with the corresponding activation from the clean run, and then measure the change of the output towards the correct answer. This is called \textit{denoising}. Alternatively, in \textit{noising}, one patches from the corrupted run into the clean run; cf. Figure \ref{fig:activation}.   Through trying many different activations and assessing how much they affect the corrupted run, one can identify the activations that really matter. 
    Finally, we point out that patching into a transformer can be done in a number of different ways, for example, into MLP layers, attention heads, or the values of the residual stream. 
    We refer the reader to \cite[chap. 1.3 \S3]{ARENA} for more details.
    \item[$\blacktriangleright$] The activation patching approach considers the alternative of swapping the contribution of an attention head to the residual stream with what it would have been under a different distribution while everything else remains the same. On the other hand, in path patching, one contemplates the alternative of replacing the input to an attention head from another attention head with what it would have been under a different distribution; see Figure \ref{fig:path}. The path-patching tool, thus, investigates the importance of particular paths between a model's components (e.g., a circuit formed by two attention heads) rather than individual model components (e.g., a single attention head).  Again, a clean run and a corrupted run are involved. Looking at the IOI task again, instead of replacing \textsf{''When Mary and John went to the store, John gave a drink to ...''} with \textsf{''When Mary and John went to the store, Mary gave a drink to ...''}, three random names are used as in \textsf{''When [X] and [Y] went to the store, [Z] gave a drink to ...''}---hence the information about duplicate tokens is erased. We refer the reader to \cite[chap. 1.3 \S4]{ARENA} for more details.
\end{itemize}
\item How does one assess if a discovered circuit is a reliable explanation for the model behavior? Paper \cite{wang2022interpretability} proposes three evaluation criteria for circuit analysis:
\begin{itemize}
    \item Faithfulness: can the circuit perform the task as well as the whole model?
    \item Completeness: does the circuit contain all the nodes used to perform the task?
    \item Minimality: are all the nodes in the circuit relevant to the task?
\end{itemize}
\end{itemize}

\begin{figure}
\includegraphics[width=14.5cm]{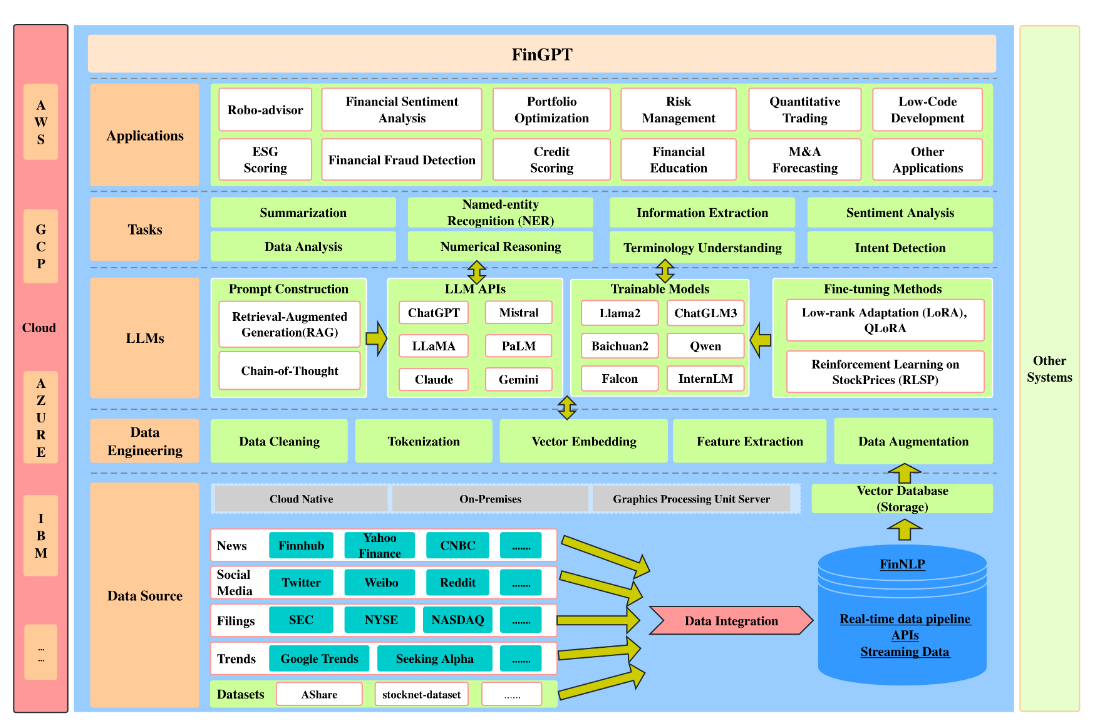}
\caption{An illustration of the FinGPT ecosystem. Picture adapted from \url{https://github.com/AI4Finance-Foundation/FinGPT} (under the MIT license).}
\label{fig:Fin}
\end{figure}




\section{Applications in financial services}
\label{s3}
After the remarkable success of ChatGPT, there have been efforts to design domain-specific large language models tailored to specific sectors such as law, commerce, healthcare and finance (see \cite{li2023large}). Such models are constructed through pre-training and/or fine-tuning on targeted datasets to perform well-defined tasks specific to a particular domain. They can tackle data security issues and prevent AI hallucination in specialized fields \cite{LLMSpecific}.

\subsection{LLM interpretability for financial services}
\label{section31}
In the financial domain, BloombergGPT  is an LLM with 50 billion parameters trained on a wide range of financial data \cite{2023arXiv230317564W}. BloombergGPT is not open source, and thus, investigating it through the lens of mechanistic interpretability, e.g., circuit discovery, is not practical. 
Moreover, frequently retraining an LLM model like BloombergGPT is costly. Therefore, there is a need for lightweight adaptations. One remarkable adaptation is the \href{https://github.com/AI4Finance-Foundation/FinGPT}{FinGPT} project which facilitates \textit{prompt engineering} and fine-tuning of open-source LLMs on financial data \cite{2023finnlp} for a variety of financial applications such as risk management, portfolio optimization, credit scoring, and fraud detection.

In the financial services industry, LLMs, and, more generally, NLP models, have been used for various customer-service related tasks. In what follows, we outline a few industry-wide compliance, legal and business use cases.\footnote{Disclaimer: The authors have neither developed/fine-tuned any large language model for these use cases nor have they used any customer data or internal data from Discover Financial Services for their experiments.} 

\begin{itemize}

\item An existing large language model, such as BERT, GPT-2, or LLaMA-2,  may be utilized, through fine-tuning, prompt engineering or \textit{Retrieval Augmented Generation} (RAG) for in-house legal and compliance applications such as processing legal documents and correspondence, conducting sentiment analysis to monitor compliance with (or potential violations of) a variety of regulations overseeing financial services. For instance: 

\begin{itemize}
\item Monitoring cases and complaints involving Unfair or Deceptive Acts or Practices (UDAAP) \cite{UDAAP};
\item Detecting cases related to Telephone Consumer Protection Act (TCPA) \cite{TCPA}, e.g., circumstances where customers requests involve TCPA regulations;
\item Compliance monitoring for regulations concerning military lending protections (e.g., Military Lending Act (MLA) \cite{MLA} and Servicemembers Civil Relief Act (SCRA) \cite{SCRA});
\item Compliance monitoring for consumer privacy protection (as required by the California Consumer Privacy Act (CCPA) \cite{CCPA});
\item Compliance monitoring for fair lending (FL) laws (as required by the Equal Credit Opportunity Act (ECOA) \cite{ECOA} and Fair Housing Act (FHA) \cite{FHA}).
\end{itemize}
\item An existing large language model can be used (through fine tuning, for example)  to classify customer-agent transcripts with business-related tags such as ``website/mobile'', ``login issues'', ``hard inquiry", ``balance too high'' etc.      
\end{itemize}

Now, consider the problem of explaining an LLM model applied to a regulatory, legal, or business use case in financial services. Inspired by \cite{wang2022interpretability}, the approach we adapt here is to design algorithmic tasks. 
In what follows, we describe how financial algorithmic tasks can be designed by working on some examples. These tasks can often be decomposed into a set of rules. These rulesets are based on legal requirements and business criteria and are often handcrafted by subject-matter experts. Below, we describe these rules for some of the compliance tasks described above followed by some algorithmic examples that can be constructed. 
\begin{itemize}
\setlength\itemsep{1em}
    \item \textbf{TCPA} rules) The TCPA regulation \cite{TCPA} prohibits unwanted communication with consumers. In compliance with this, certain conversations should be classified as critical governed by TCPA.
    One may design the following algorithmic examples for TCPA use cases in order to generate datasets to be used for mechanistic interpretability investigations:
    \begin{itemize}
        \item[$\blacktriangleright$] MARKETING-CALL-TEMPLATES: \textsf{''The agent reaches out to the customer regarding a new [FINANCIAL-PRODUCT]. The customer says I'm not interested. Please remove me from your call list.''}---\textsf{[FINANCIAL-PRODUCT]} consists of: \textsf{credit card}, \textsf{auto loan}, \textsf{mortgage loan}, \textsf{checking account}, \textsf{savings account}.
        
        \item[$\blacktriangleright$] COMMUNICATION-TEMPLATES: \textsf{''I don't want to receive any [WAY-of-COMMUNICATIONS] from you.''}---\textsf{[WAY-of-COMMUNICATIONS]} is given by: \textsf{calls}, \textsf{mails}, \textsf{text messages}, \textsf{messages}, \textsf{emails}, \textsf{notifications}, \textsf{communications}, \textsf{further communications}, etc.

        \item[$\blacktriangleright$] STOP-CONTACT-LIST-TEMPLATE: \textsf{``Please add my [PROFILE] to the do-not-call lists. I don't want to be contacted anymore.''}---\textsf{[PROFILE]} consists of the customer contact details such as: \textsf{number}, \textsf{phone number}, \textsf{personal number}, \textsf{work number}, \textsf{email address}, \textsf{personal email}, \textsf{mail address}, etc.

        \item[$\blacktriangleright$] DEBT-COLLECTION-CALL-TEMPLATE: \textsf{``The agent notifies the customer regarding an \newline[OVERDUE-BALANCE] on her account and offers a payment plan. The customer says I don't want to receive calls about this anymore.''}---\textsf{[OVERDUE-BALANCE]} is given by: \textsf{missed payment}, \textsf{missed minimum payment}, \textsf{outstanding balance}, \textsf{overdue balance}, \textsf{unpaid balance}, \textsf{delinquent balance}, etc.
    \end{itemize}
    \item \textbf{UDAAP} rules) UDAAP protects consumers against the risks of harm from unfair, deceptive, or abusive practices by financial institutions. Harm does not have to be monetary and can result from increased difficulty of consumer understanding of the overall costs or risks of the product and the potential harm to the consumer associated with the product. Financial institutions monitor their customer-agent interactions. When the consumer suggests potential UDAAP (Unfair, Deceptive, or Abusive Acts or Practices) concerns, such interactions need to be identified for further review by a financial institution. One may use the following criteria to define UDAAP as an algorithmic financial task:
    \begin{itemize}
        \item UDAAP Keywords: \textsf{Unfair/not fair}, \textsf{deceptive/deceitful}, \textsf{abusive/abuse/abused}, \textsf{tricky/trick/tricked}, \textsf{cheating/cheat/cheated},   \textsf{misleading/mislead/misled}, \textsf{lying/lie/lied}, \textsf{taking/took advantage of me because I'm a [member of a vulnerable population]} (e.g., students and minority groups).
        \end{itemize}
        Therefore, we design the following algorithmic examples for UDAAP circuit discovery:
        \begin{itemize}
        \item[$\blacktriangleright$] UNFAIR-PRACTICES: The customer is interested in opening a checking account and asks about overdraft protection. The agent mentions that the account comes with overdraft protection but fails to clearly disclose the high fees associated with overdraft transactions. Example: \textsf{''You are [UDAAP-BEHAVIOR]. You did not disclose the high fees associated with your \textsf{[FINANCIAL-PRODUCT]}.''}---\textsf{[UDAAP-BEHAVIOR]} can be replaced with the followings: \textsf{unfair}, \textsf{deceptive}, \textsf{deceitful}, \textsf{abusive}, \textsf{misleading}, \textsf{fraud}, \textsf{liars} etc. 
        \item[$\blacktriangleright$] DECEPTIVE-PRACTICES: The customer is interested in a personal loan and the agent mentions that they offer personal loans with APR as low as $4.99\%$. Example: \textsf{''You are [UDAAP-BEHAVIOR]. You did not disclose that the rate $4.99\%$ only applies to applicants with excellent credit scores and my rate is significantly higher. Your loans also come with hidden origination fees that the agent did not mention."}
        \item[$\blacktriangleright$] ABUSIVE-PRACTICES: The agent engages in aggressive debt collection by notifying the customer that he is 30 days overdue on his credit card payment and he must immediately pay his debt, or the company will take legal actions against him. Example: \textsf{"You are [UDAAP-BEHAVIOR]. You are asking me to pay the full amount now and are threatening me that you will garnish my  wages and reporting me to credit bureaus."}
        \end{itemize}
    \item \textbf{Fair Lending} rules) These laws set forth prohibited bases (i.e., protected classes) against which a financial institution cannot discriminate.  
    \begin{itemize}
        \item Protected classes:
        \begin{itemize}
        \item Race, national origin, age, disability, religion, sex, marital or familial status (i.e., having children).
        \item An applicant’s receipt of income from public assistance or their good faith exercise of any right under the Consumer Credit Protection Act. 
        \item States may have broader protected classes under their respective fair lending laws, including but not limited to military status, ancestry, domestic violence victim status, or political affiliation.
        \end{itemize}
        \item Examples of Potential Discrimination: 
        \begin{itemize}
        \item \textsf{''I requested for a credit line increase multiple times, and you said no because I'm on unemployment.''}
        \item \textsf{''Why are you denying all my requests for a payment plan. It's because I'm a single mother.''}
        \item \textsf{''I don't understand why I have all these late and hidden fees. I think you're trying to take advantage of me because I'm disabled.''}
        \item \textsf{''You are quoting me a much higher rate than my neighbor when we have the same credit score and income. It's because I'm old!''}
        \end{itemize}
        \end{itemize}
        We design the following algorithmic examples for Fair Lending circuit discovery:
        \begin{itemize}
        \item[$\blacktriangleright$]
        \textsf{''I don't understand why I have all these late fees. I think you're trying to [UDAAP-VERB] me because I'm a [MEMBER-of-PROTECTED-CLASS].''}---\textsf{[UDAAP-VERB]} is given by: \textsf{trick}, \textsf{cheat}, \textsf{deceit}, \textsf{abuse}, \textsf{mislead}, \textsf{defraud}, \textsf{taking advantage of}. 
        \item[$\blacktriangleright$] Note that \textsf{[MEMBER-of-PROTECTED-CLASS]} can be an item from \textsf{elderly}, \textsf{single mother}, \textsf{disabled}, \textsf{minority}, etc.
        \begin{itemize}
        \item \textsf{''You are taking advantage of me because I'm a [MEMBER-of-PROTECTED-CLASS];''}    
        \item \textsf{''You are charging me a much higher rate than my neighbor. It's because I'm a [MEMBER-of-PROTECTED-CLASS];''}
        \item \textsf{''I asked for a credit line increase, and you denied my request because I'm a [MEMBER-of-PROTECTED-CLASS].''}
        \end{itemize}
        \item[$\blacktriangleright$] The customer is looking for a loan to expand their small business. The agent provides less favorable loan terms to the business. Example: \textsf{"You are providing us with a less favorable loan terms compared to other similar businesses with similar business plans and finances. Is this because we are a [MINORITY-OWNED] business?"}---\textsf{[MINORITY-OWNED]} can be given by: \textsf{women-owned}, \textsf{black-owned}, \textsf{LGBTQ+-owned}, etc.
        \end{itemize}           
\end{itemize}

\subsection{Numerical Experiments} In this section, we design mechanistic interpretability examples for compliance tasks in financial services. For the sake of simplicity, we choose GPT-2 Small \cite{radford2019language} as the model to study mechanistically following the methods presented in \cite{wang2022interpretability}. For the rest of this section, we adapt the steps presented in \cite{ARENA} for the IOI task and tailor it to the compliance tasks described in \S\ref{section31}.

We focus on understanding financial language models tasks involving Fair Lending laws discussed before. Our goal is to analyze GPT-2 Small attention pattern in identifying potential violation of Fair Lending Laws related to the Equal Credit Opportunity Act (ECOA) \cite{ECOA} and the Fair Housing Act (FHA) \cite{FHA}. In doing so, we design prompts of the following format: \textsf{"The customer says on the phone that you are denying my request for a payment plan because I'm on unemployment."}; and then we ask \textsf{"Is this is an example of a Fair Lending violation based on Equal Credit Opportunity Act (ECOA)?"}. Next, we measure model performance by defining the following Logit Difference\footnote{Note that this choice of metric was inspired by \cite{wang2022interpretability} where they study the logit difference between the indirect object and the subject tokens for the circuit discovery of the IOI task.} metric: $logit$(\textsf{"Yes"}) $-$ $logit$(\textsf{"No"}). Furthermore, we define another metric where we compare the \textsf{"Yes"} and \textsf{"No"} token probabilities of the final output, i.e., $P(\textsf{"Yes"})/P(\textsf{"No"})$. The prompts that we investigate are illustrated in Table 2.
\begin{figure*}
\includegraphics[height=9cm, trim={0 0 0 1cm},clip]{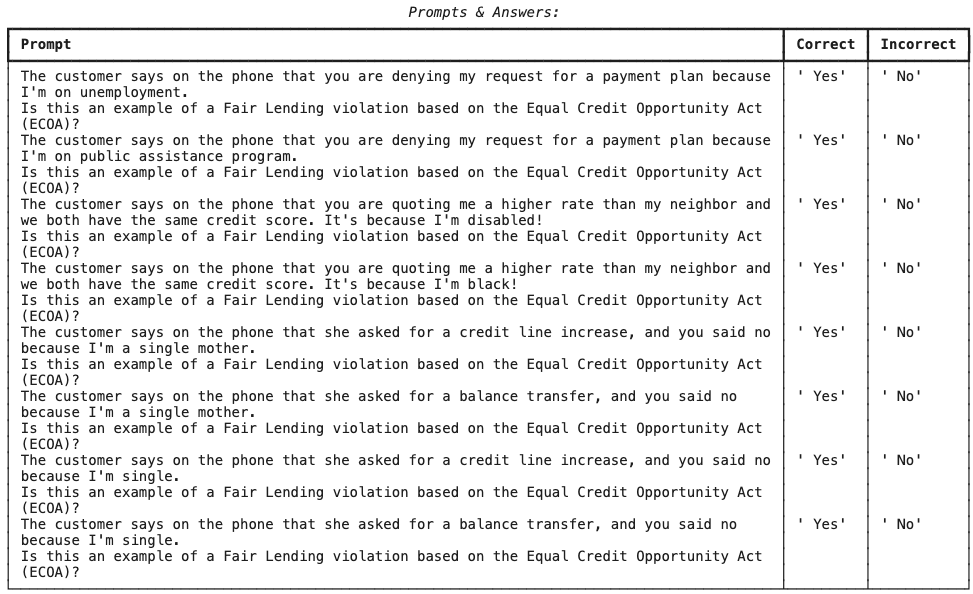}
\caption*{Table 2: An illustration of the Fair Lending rule financial task investigation. Prompts and the answers.}
\label{fig:Prompt_Answer}
\end{figure*}
The corresponding logit differences and normalized probability ratios for the prompts are presented in Table 3. The average logit difference and probability ratio are $0.81$ and $2.26$, respectively.
\begin{figure*}
\includegraphics[height=3.2cm, trim={0 0 0 1cm},clip]{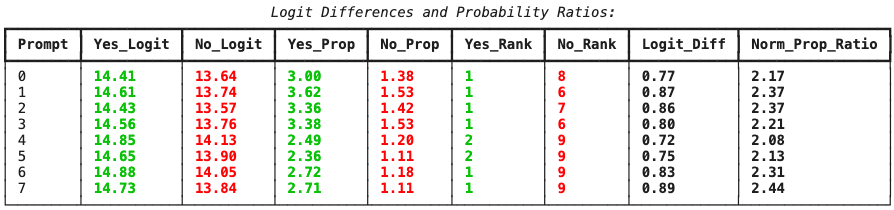}
\caption*{Table 3: Final token logits, token probabilities, and token ranks for the \textsf{"Yes"} and \textsf{"No"} tokens, along with the logit differences and the normalized probability ratios.}
\label{fig:num_res}
\end{figure*}

Next, we study the model by performing direct logit attribution \cite{wang2022interpretability}. More specifically, we investigate the contributions of each layer (and its corresponding attention heads) to the logit difference of the residual stream.   

\begin{figure}
\includegraphics[height=8cm, trim={0 1cm 0 3cm},clip]{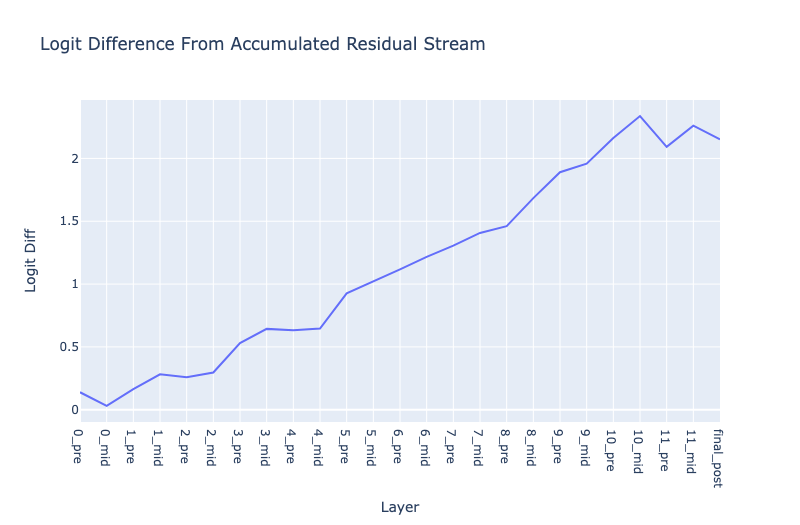}
\caption{Calculated logit difference for the decomposed accumulated residual stream after each layer. \textsf{n-pre} denotes the residual stream at the start of layer \textsf{n}, while \textsf{n-mid} denotes the residual stream after the attention part of layer \textsf{n}.}
\label{fig:newplot}
\end{figure}

Figure \ref{fig:newplot}, illustrates the calculated logit difference by decomposing the residual stream after each layer (see \emph{logit lens} from \cite{nanda2022transformerlens}).\footnote{Recall from \S\ref{section23} that, one can interpret the result of residual stream decomposition and calculating logit differences as simulating what would happen if subsequent layers are removed after each layer \cite{ARENA}.} As one can see the logit difference generally keeps improving through the layers. This is in contrast to the IOI task where one can localize the task completion to merely a few final layers, i.e., layers $7$, $8$, and $9$ \cite{ARENA}. We believe that this is due to the fact that our compliance task at hand is far more complex when compared to the IOI task which involves moving information around from the indirect object and not the subject, requiring less information processing.   

\begin{figure}
\includegraphics[height=8cm, trim={0 1cm 0 3cm},clip]{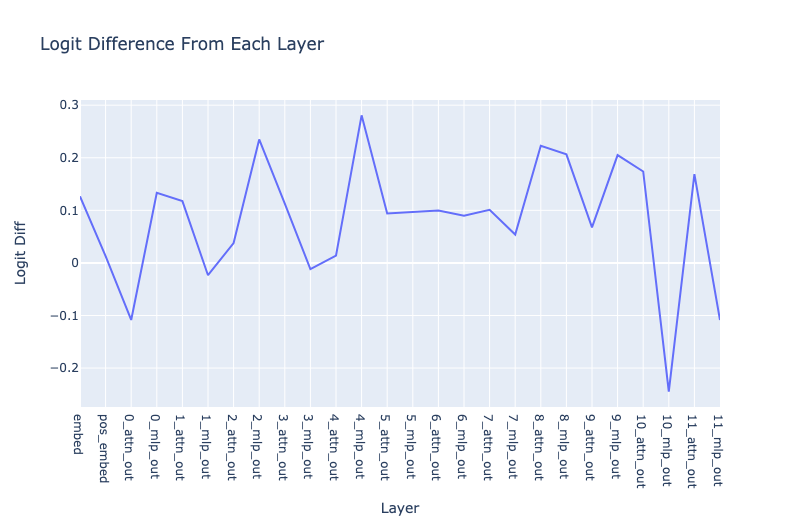}
\caption{Break down of logit differences from each layer between adjacent residual streams.}
\label{fig:newplot1}
\end{figure}

We may investigate the logit difference between adjacent residual streams (see Figure \ref{fig:newplot1}). One can realize that unlike the IOI task, where only attention layers matter, MLP layers play a significant role in performing the financial compliance task. In particular, MLP layers $2$ and $4$ improve task completion greatly, while MLP layer $10$ and attention layer $0$ decrease performance.

\begin{figure}
\includegraphics[height=6.5cm, trim={0 0 0 3cm},clip]{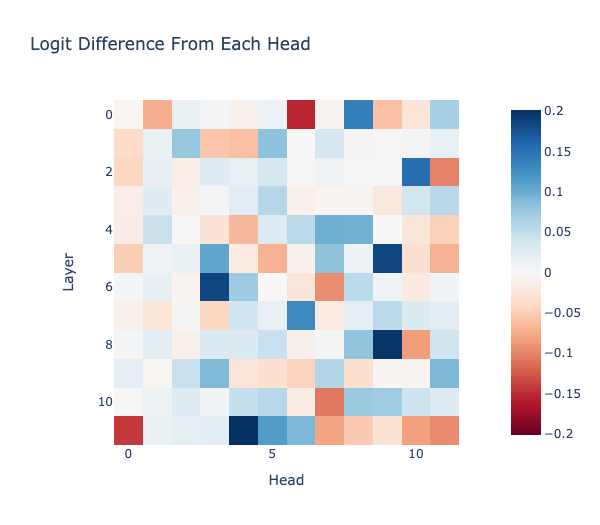}
\caption{Attention heads heat map per layer illustrating logit difference from each head.}
\label{fig:newplot2}
\end{figure}

\begin{figure}
\includegraphics[height=10.3cm, trim={0 0 29.5cm 0},clip]{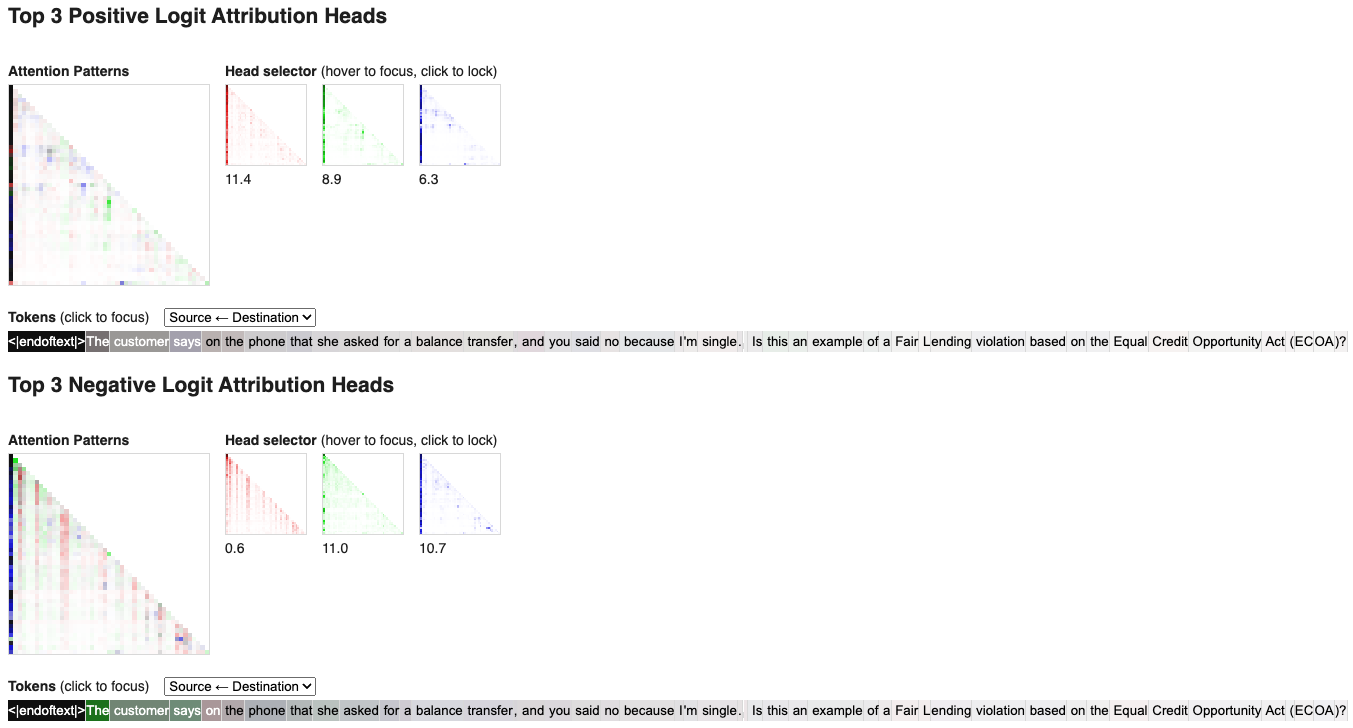}
\caption{Attention patterns for the top three positive and top three negative logit attribution heads.} 
\label{fig:newplot3}
\end{figure}

Next, we further decompose the output of each attention layer into the sum of the outputs of the corresponding attention heads (see Figure \ref{fig:newplot2}). As expected, there are quite a few attention heads that play a substantial role in the compliance financial task completion. Of note, heads\footnote{Note we use the convention that \textsf{l.h} specifies the head number \textsf{h} in the layer number \textsf{l} (indexing starts from $0$). There are $144 (12\times 12)$ heads in total in GPT-2 Small.} 11.4, 8.9, and 6.3 contribute a lot positively, while heads 0.6, 11.0, and 10.7 contribute a lot negatively to the task completion (see Figure \ref{fig:newplot3}).

\begin{figure}
\includegraphics[height=5cm, trim={0 0 0 0.9cm},clip]{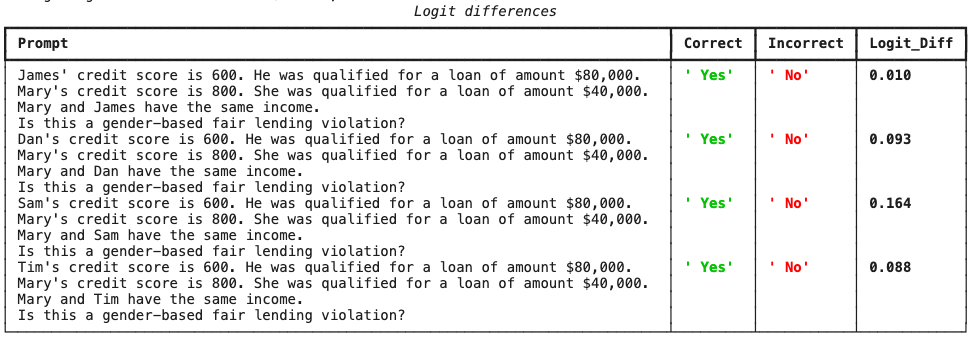}
\caption{Logit difference for \textsf{"Yes"} (ranked 1st) and \textsf{"No"} (ranked 2nd) answer tokens for the clean prompts.} 
\label{fig:act1}
\end{figure}

\begin{figure}
\includegraphics[height=5cm, trim={0 0 0 0.9cm},clip]{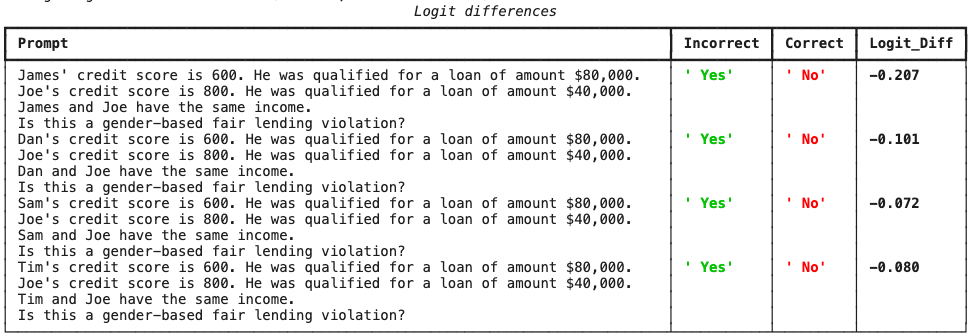}
\caption{Logit difference for \textsf{"Yes"} (ranked 2nd) and \textsf{"No"} (ranked 1st) answer tokens for the corrupted prompts.} 
\label{fig:act2}
\end{figure}

As mentioned in \S\ref{section23}, one of the limitations of the direct logit attribution discussed before is that it only looks at the very end of the circuit which affects the logits directly \cite{ARENA,wang2022interpretability}. To obtain a more refined understanding, in what follows, we utilize the activation patching technique. In doing so, we design the so-called clean prompts (Figure \ref{fig:act1}), and corrupted prompts (Figure \ref{fig:act2}). 
The format for the clean prompts is given by: \textsf{"[A]'s credit score is 600. He was qualified for a loan of amount $\$80,000$. [B]'s credit score is 800. She was qualified for a loan of amount $\$40,000$. [B] and [A] have the same income. Is this a gender-based fair lending violation?"}, where [A] and [B] are, respectively, predominately male-associated and female-associated names. Similarly, the corrupted prompt is given by: \textsf{"[A]'s credit score is 600. He was qualified for a loan of amount $\$80,000$. [C]'s credit score is 800. He was qualified for a loan of amount $\$40,000$. [A] and [C] have the same income. Is this a gender-based fair lending violation?"}, where [A] and [C] are predominately male-associated names.
The clean run favors \textsf{"Yes"} over \textsf{"No"} as an output response, and thus, results in positive logit differences, while the corrupted run favors \textsf{"No"} over \textsf{"Yes"} as an output response resulting in negative logit differences. Next, we run the model on our corrupted prompts and then intervene by patching in the activations from our clean prompts and monitor how the output response changes. 


\begin{figure}
\includegraphics[height=6.7cm, trim={0 1cm 0 3cm},clip]{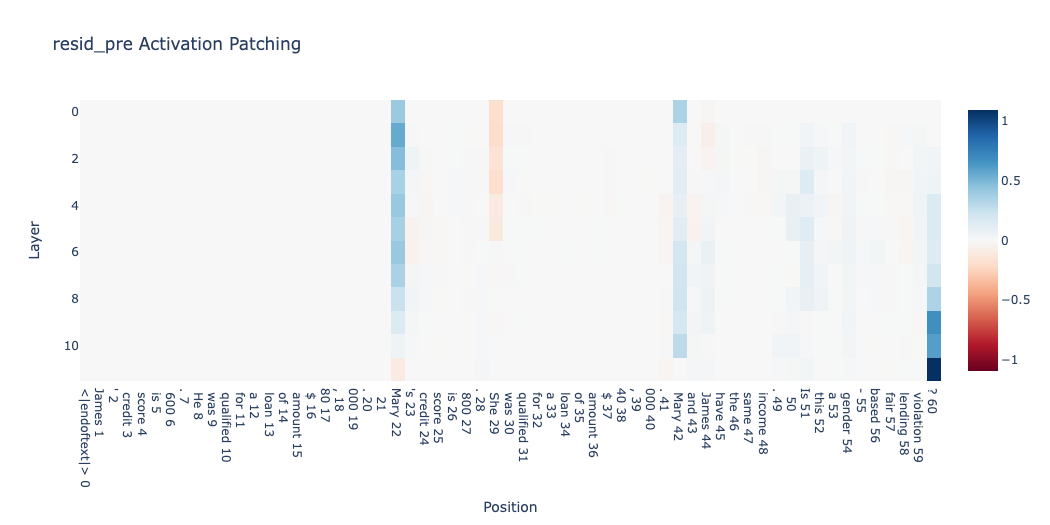}
\caption{Residual stream patching at the onset of each layer for each token positions averaged over all $4$ prompts. On the x-axis we only show the labels for the first prompt.} 
\label{fig:act3}
\end{figure}

Figure \ref{fig:act3} depicts the results for residual stream patching at the start of each layer for all the $61$ token positions. A score closer to $0$ means the performance is closer to the one obtained on the corrupted input, while a score closer to $1$ means that the performance is closer to that of the clean input. We are trying to look for activations that are sufficient to recover the correct response. One can make the following observations: (i) The computation is fairly localized to positions: $22$ (\textsf{"[B]"} in "\textsf{[B]'s credit score ..."}), $29$ (\textsf{"She"} in \textsf{"She was qualified ..."}), $42$ (\textsf{"[B]"} in \textsf{"[B] and [A] ..."}), $44$ ("[A]" in \textsf{"[B] and [A] ..."}), $54$ (\textsf{"gender"} in \textsf{"gender-based ..."}), and \textsf{END} tokens. (ii) One can see that the information at token $22$ starts to be moved to \textsf{END} token around layers $5$ and $6$.

\begin{figure}
\includegraphics[height=4.7cm, trim={0 1cm 0 3cm},clip]{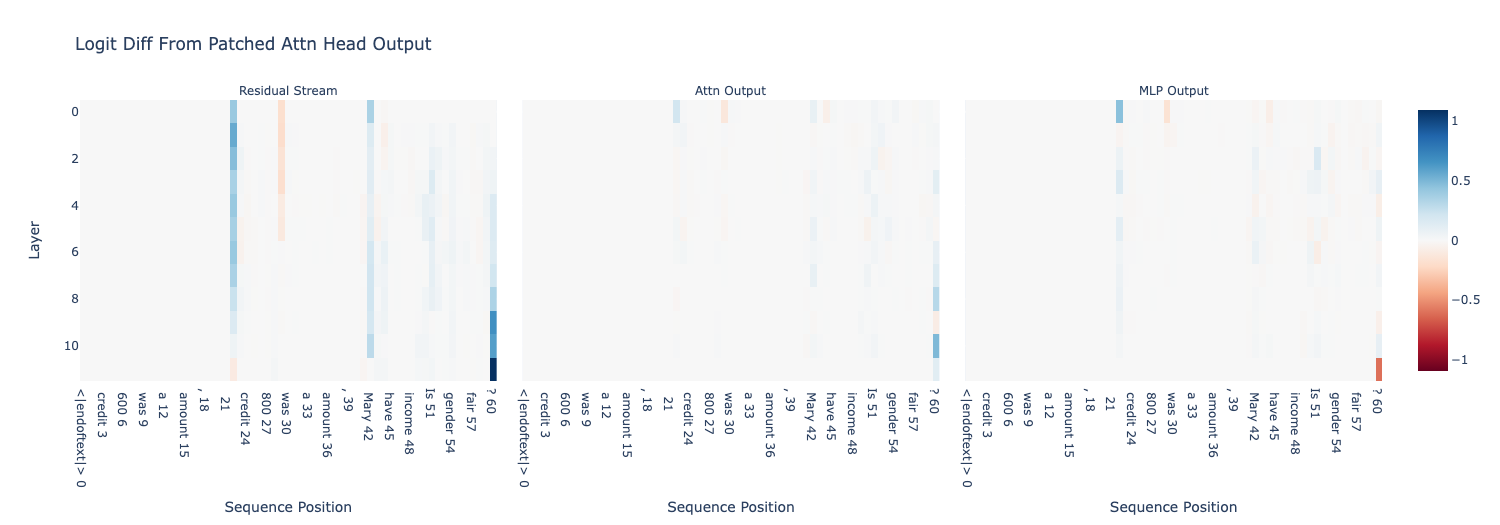}
\caption{Patching in residual stream after the attention layer (in the middle) and after the MLP layer (on the right).}
\label{fig:act5}
\end{figure}

Instead of just patching to the residual stream at the onset of each layer, we may use activation patching for patching after the attention layer and after the MLP layer. Figure \ref{fig:act5} illustrates the results. As one can see layers $8$ and $10$ contribute positively, while layer $9$ contributes negatively to the performance. Among MLP layers, we observe that \textsf{MLP-0}\footnote{Note that this is consistent with the observations from \cite{wang2022interpretability, ARENA}, that \textsf{MLP-0} matters a lot. This behavior is often observed in GPT-2 Small.} plays an important role, along with \textsf{MLP-11}. 

\begin{figure}
\includegraphics[height=6.5cm, trim={0 1.2cm 0 3cm},clip]{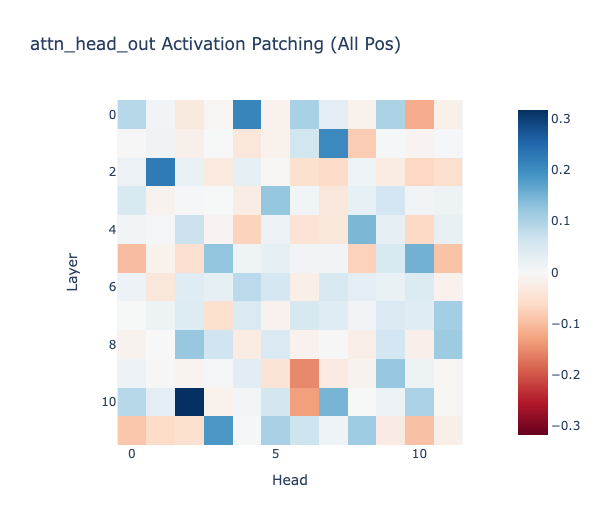}
\caption{Patching in individual attention heads output for all layers and sequence positions.} 
\label{fig:act6}
\end{figure}

Next, we can further refine our analysis (see Figure \ref{fig:act6}) by patching in on individual attention heads for all layers and tokens. We see that the heads $10.2$, $10.7$, and $11.3$, from later layers, and the heads $0.4$, $1.7$, and $2.1$, from earlier layers, have very large positive scores. On the other hand, the heads $9.6$ and $10.6$, from later layers, and the heads $0.10$ and $5.0$, from earlier layers, have very large negative scores.

\begin{figure}
\includegraphics[height=3.7cm, trim={0 2cm 0 3cm},clip]{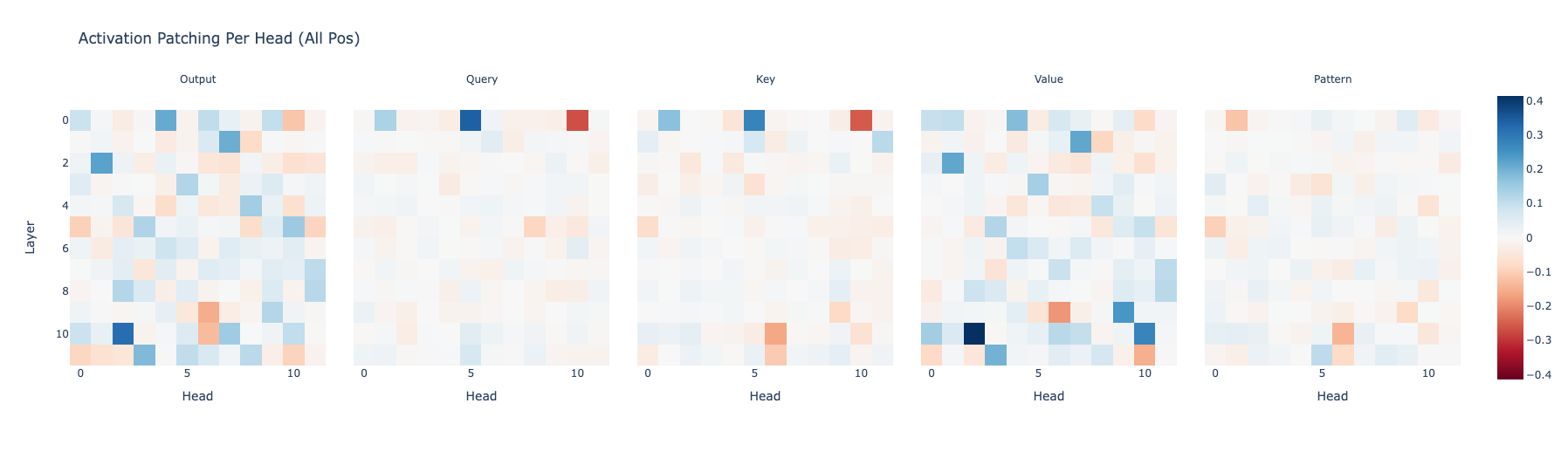}
\caption{Decomposing heads by patching on the query vectors, key vectors and value vectors.} 
\label{fig:act8}
\end{figure}

Finally, instead of just patching on a head's output, we decompose the attention heads by patching into the building blocks of the attention mechanism, i.e., key vectors, query vectors, value vectors, and attention patterns (see Figure \ref{fig:act8}). We can deduce the following observations: (i) Some of the layer-$0$ heads such as $0.5$, $0.1$ and $0.10$ are very important because of their query and key vectors. (ii) The head $10.2$ is significantly important due to its value vector. (iii) Other important attention heads owing to their value vectors are $9.6$, $9.9$, $10.10$, $11.3$, and $11.10$. (iv) Thus, we can conclude that value patching has a more significant effect than key (or query) patching for the important heads in later layers, i.e., layers $9-11$.

In summary, we investigated GPT-2 Small's attention pattern for identifying potential violation of Fair Lending laws. Employing the direct logit attribution technique, we analyzed the contributions of each layer to the logit difference in the residual stream. In particular, we identified the heads $11.4$, $8.9$, and $6.3$ ($0.6$, $11.0$, and $10.7$) that contribute a lot positively (negatively) to the compliance task completion. We performed a casual intervention through utilizing the activation patching technique by designing the clean and corrupted prompts. We refined our analysis to the find which components of the attention mechanism are important for each attention head. Of note, we found that the head $10.2$ is very important due to its value vector, while some of the layer-$0$ heads such as $0.5$, $0.1$, and $0.10$ are very important owing to their query and key vectors.

\section{Future Directions}
We consider this paper to be the first step towards leveraging mechanistic interpretability techniques for understanding applications of LLMs in financial services. Some of the future research directions include: (i) Conducting experiments with more powerful open-source LLMs such as Mistral 7B and Llama-2 7B. This is where one may leverage fine-tuned open-source models from FinGPT project \cite{2023finnlp}. (ii) Working on more algorithmic examples of compliance tasks in financial services and preferably finding prompts that result in large logit differences. (iii) Perform path patching as a more sophisticated circuit discovery approach toward finding an end-to-end circuit for performing a compliance financial task.



\section*{Acknowledgement}
We would like to thank Sharon O'Shea Greenbach  (Sr. Counsel \& Director, Regulatory Policy at DFS) and Dennis Lee (Chief IP Counsel \& AGC Privacy \& Security at DFS) for their helpful comments relevant to regulatory issues that arise in the financial industry and for carefully reviewing the manuscript. We also thank Callum McDougall, Shervin Minaee, and Christopher Olah for granting permission to use some of the figures and tables from their papers. Ashkan Golgoon benefited from stimulating discussions with Amirhossein Tajdini. The views and opinions expressed in this paper are solely our own and do not represent or reflect those of our employer.

\bibliography{bibliography}
\bibliographystyle{abbrv}

\end{document}